\documentclass[letterpaper, 10 pt, conference]{ieeeconf}  

\IEEEoverridecommandlockouts                              

\usepackage{amsmath}
\usepackage{amssymb}
\usepackage{graphicx}
\usepackage{subcaption}
\usepackage{xcolor}
\usepackage{algorithm}
\usepackage{algpseudocode}
\usepackage{cite}
\usepackage{subcaption}
\usepackage{url}

\newtheorem{definition}{Definition}
\newtheorem{theorem}{Theorem}

\newtheorem{problem}{Problem}
\newtheorem{proposition}{Proposition}

\newtheorem{assumption}{Assumption}

\title{\LARGE \bf
Fault Tolerant Neural Control Barrier Functions for Robotic Systems under Sensor Faults and Attacks
}

\author{Hongchao Zhang$^{1}$, Luyao Niu$^{2}$, Andrew Clark$^{1}$, and Radha Poovendran$^{2}$
\thanks{$^{1}$Hongchao Zhang and Andrew Clark are with the Electrical and Systems Engineering Department, McKelvey School of Engineering, Washington
University in St. Louis, St. Louis, MO 63130
        {\tt\small \{hongchao, andrewclark\}@wustl.edu}}%
\thanks{$^{2}$Luyao Niu and Radha Poovendran are with the Network Security Lab, Department of Electrical and Computer Engineering,
University of Washington, Seattle, WA 98195-2500
        {\tt\small \{luyaoniu,rp3\}@uw.edu}}%
}

\begin{document}

\maketitle
\thispagestyle{empty}
\pagestyle{empty}

\begin{abstract}

Safety is a fundamental requirement of many robotic systems. 
Control barrier function (CBF)-based approaches have been proposed to guarantee the safety of robotic systems.
However, the effectiveness of these approaches highly relies on the choice of CBFs.
Inspired by the universal approximation power of neural networks, there is a growing trend toward representing CBFs using neural networks, leading to the notion of neural CBFs (NCBFs).
Current NCBFs, however, are trained and deployed in benign environments, making them ineffective for scenarios where robotic systems experience sensor faults and attacks.
In this paper, we study safety-critical control synthesis for robotic systems under sensor faults and attacks.
Our main contribution is the development and synthesis of a new class of CBFs that we term fault tolerant neural control barrier function (FT-NCBF).
We derive the necessary and sufficient conditions for FT-NCBFs to guarantee safety, and develop a data-driven method to learn FT-NCBFs by minimizing a loss function constructed using the derived conditions.
Using the learned FT-NCBF, we synthesize a control input and formally prove the safety guarantee provided by our approach.
We demonstrate our proposed approach using two case studies: obstacle avoidance problem for an autonomous mobile robot and spacecraft rendezvous problem, with code available via \url{https://github.com/HongchaoZhang-HZ/FTNCBF}.
\end{abstract}

\section{INTRODUCTION}

Robotic systems are increasingly deployed in safety-critical applications such as search and rescue in hazardous environments \cite{fernandez2023distributed,peng2023safe,jian2023dynamic}. 
Safety requirements are normally formulated as the positive invariance of given regions in the state space.
Violations of safety could lead to catastrophic damage to robots, harm to humans that co-exist in the field of activities, and economic loss \cite{knight2002safety,schwarting2018planning}.
A popular class of methods for safety-critical synthesis is control barrier function (CBF)-based approaches \cite{ames2019control,xu2015robustness,Wei2021high,tan2021high}.
A CBF defines a positive invariant set within the safety region such that when the robot reaches the boundary of the set, the control input will steer the robot towards the interior of the set.

The performance and safety guarantees of CBF-based approaches are strongly dependent on the choice of barrier functions.
Neural control barrier functions (NCBFs) \cite{dawson2022safe,dawson2023safe,liu2023safe}, which represent CBFs using neural networks, have attracted increasing interest.
Compared with the polynomial CBFs found by sum-of-squares (SOS) optimization \cite{clark2021verification,clark2022semi,kang2023verification,dai2022convex,jagtap2020formal}, 
NCBFs leverage the universal approximation power of neural networks \cite{tabuada2022universal,hornik1989multilayer,csaji2001approximation}, and thus allow CBFs to be applied to high-dimensional complex systems \cite{srinivasan2020synthesis,xiao2021barriernet,dawson2023safe,dawson2022learning,robey2020learning,lindemann2021learning}, e.g., learning-enabled robotic systems \cite{tampuu2020survey} and neural network dynamical models \cite{raissi2019physics,gamboa2017deep}.



At present, NCBFs \cite{dawson2022safe,dawson2023safe,liu2023safe, mathiesen2022safety, mazouz2022safety, zhao2020synthesizing} are developed for robotic systems designed to operate in fault- and attack-free environments.
Sensors mounted on robotic systems, however, have been shown to be vulnerable to a wide range of faults and malicious attacks \cite{mo2010false,guan2017distributed}.
As demonstrated in our experiments, the safety guarantees of NCBFs do not hold when robots are operated in such adversarial environments.

In this paper, we study the problem of safety-critical control of robotic systems under sensor faults and attacks.
We consider that an adversary can choose an arbitrary fault or attack pattern among finitely many choices, where each pattern corresponds to a distinct subset of compromised sensors.
We propose a new class of CBFs called fault-tolerant NCBFs (FT-NCBFs).
We present a data-driven method to learn FT-NCBFs.
Given the learned FT-NCBFs, we then formulate a quadratic program to compute control inputs.
The obtained control inputs guarantee that the robot moves towards the interior of the positive invariant set defined by the FT-NCBF under all attack patterns.
Consequently, we can guarantee robot's safety by ensuring that the positive invariant set is contained within the safety region.
To summarize, this paper makes the following contributions.
\begin{itemize}
    \item We propose FT-NCBFs for robotic systems under sensor faults and attacks. We derive the necessary and sufficient conditions for FT-NCBFs to guarantee safety.
    Based on the derived conditions, we develop a data-driven method to learn FT-NCBFs.
    \item We develop a fault-tolerant framework which utilizes our proposed FT-NCBFs for safety-critical control synthesis. We prove that the synthesized control inputs guarantee safety under all fault and attack patterns.
    \item We evaluate our approach using two case studies on the obstacle avoidance problem of a mobile robot and the spacecraft rendezvous problem. We show that our approach guarantees the robot to satisfy the safety constraint regardless of the faults and attacks, whereas the baseline employing the existing NCBFs fails.
\end{itemize}

The rest of this paper is organized as follows. 
Section \ref{sec:formulation} presents preliminaries and problem formulation.
We present our solution in Section \ref{sec:sol} and demonstrate its safety guarantee in Section \ref{sec:exp}.
Section \ref{sec:conclusion} concludes the paper.





\section{Preliminaries and Problem Formulation}\label{sec:formulation}
This section presents the system and adversary models. 
We then state the problem studied in this paper.
We finally introduce preliminary background on extended Kalman filters and stochastic control barrier functions.

\subsection{System and Adversary Model} 
We consider a robotic system with state $x_{t} \in \mathcal{X}\subseteq\mathbb{R}^{n}$ and input $u_{t} \in \mathbb{R}^{p}$ at time $t$. The state dynamics and  output $y_{t} \in \mathbb{R}^{q}$ are described by the stochastic differential equations
\begin{align}
    \label{eq:state-sde}
    dx_{t} &= (f(x_{t})+g(x_{t})u_{t}) \ dt + \sigma_{t} \ dW_{t}\\
    \label{eq:output-sde}
    dy_{t} &= (c x_{t} + a_{t}) \ dt + \nu_{t} \ dV_{t}
\end{align}
where functions $f: \mathbb{R}^{n} \rightarrow \mathbb{R}^{n}$ and $g: \mathbb{R}^{n} \rightarrow \mathbb{R}^{n \times p}$ are locally Lipschitz, $\sigma_{t} \in \mathbb{R}^{n \times n}$, $W_{t}$ is an $n$-dimensional Brownian motion. In addition, matrix $c \in \mathbb{R}^{q \times n}$,  $\nu_{t} \in \mathbb{R}^{q \times q}$, and $V_{t}$ is a $q$-dimensional Brownian motion.
Here $a_{t}\in \mathbb{R}^{q}$ is an attack signal injected by an adversary, which will be detailed later in this section.

We define a control policy $\mu:\{y_{t^{\prime}} : t^{\prime} \in [0,t)\}\rightarrow \mathbb{R}^p$ to be a mapping from the sequence of outputs to a control input $u_t\in\mathbb{R}^p$ at each time $t$. 
The control policy needs to guarantee the robot to satisfy a safety constraint, which is specified as the positive invariance of a given safety region. 
\begin{definition}[Safety]
    \label{def:positive-invariance}
    A set $\mathcal{D} \subseteq \mathbb{R}^{n}$ is positive invariant under dynamics \eqref{eq:state-sde}, \eqref{eq:output-sde} and control policy $\mu$ if $x_0\in \mathcal{D}$ and $u_{t} = \mu(x_{t}) \ \forall t \geq 0$ imply that $x_t \in \mathcal{D}$ for all $t \geq 0$. 
    If $\mathcal{D}$ is positive invariant, then the system satisfies the safety constraint with respect to $\mathcal{D}$.
\end{definition}

We denote the safety region as
\begin{equation}\label{eq:safe-region}
    \mathcal{C}= \{{x} : h({x}) \geq 0\},
\end{equation} 
where $h: \mathbb{R}^{n} \rightarrow \mathbb{R}$ is locally Lipschitz.
We further let the interior and boundary of $\mathcal{C}$ be $\mbox{int}(\mathcal{C}) = \{{x} : h({x}) > 0\}$ and $\partial \mathcal{C} = \{{x} : h({x}) = 0\}$, respectively.
We assume that $x_{0} \in \mbox{int}(\mathcal{C})$, i.e., the system is initially safe. 

We consider the presence of an adversary, who can inject an arbitrary attack signal, denoted as $a_{t}\in \mathbb{R}^{q}$, to manipulate the output at each time $t$. The adversary aims to force the robot leaves the safety region $\mathcal{C}$. The attack signal is constrained by $\mbox{supp}(a_{t}) \subseteq \mathcal{F}(r)$, where $r\in\{r_1,\ldots,r_m\}$ is the index of possible faults or attacks, $m$ is the total number of possible attack patterns, and $\mathcal{F}(r)\subseteq\{1,\ldots,q\}$ denotes the set of potentially compromised sensors under attack pattern $r$.
Hence, if attack pattern $r$ occurs, then the outputs of any of the sensors in $\mathcal{F}(r_{i})$ can be arbitrarily modified. 
We consider that the set of possible faults or attacks is known, but the exact attack pattern that has occurred is unknown to the controller. 
In this paper, we make the following assumption.
\begin{assumption}
\label{assumption:sf}
The robot in Eq. \eqref{eq:state-sde}-\eqref{eq:output-sde} and the attack patterns $\mathcal{F}(r_1),\ldots,\mathcal{F}(r_m)$ satisfy the conditions: (i) The system is controllable, and (ii) For each $i,j \in \{1,\ldots,m\}$, the pair $[\frac{\partial \overline{f}}{\partial x}(x,u), \overline{c}_{i,j}]$ is uniformly detectable, where $\overline{c}_{i,j}$ is the corresponding matrix after removing sensors affected by $r_i$ and $r_j$ from matrix $c$.
\end{assumption}

We state the problem studied in this paper as follows.
\begin{problem}\label{prob:1}
    Given a safety region $\mathcal{C}$ defined in Eq. \eqref{eq:safe-region} and a parameter $\epsilon \in (0,1)$, construct a control policy $\mu$ such that, for any attack pattern $r \in \{r_{1},\ldots,r_{m}\}$, the probability $Pr(x_{t} \in \mathcal{C} \ \forall t) \geq (1-\epsilon)$ when attack pattern $r$ occurs.
\end{problem}

\subsection{Preliminaries}

The extended Kalman filter (EKF) \cite{reif2000stochastic} can be used to estimate the robot's state \cite{panigrahi2022localization, urrea2021kalman}.  We denote the state as $\hat{x}_t$ and let the updated estimate be as follows:
\begin{equation}\label{eq:ekf}
    d\hat{x}_t = (f(\hat{x}_{t})+g(\hat{x}_{t})u_{t})\ dt + K_t(dy_{t} - c\hat{x}_t),
\end{equation}
where $K_t=P_tc^TR_t^{-1}$ is the Kalman gain, $R_t = \nu_t\nu_t^T$, and $\hat{x}_t$ is the state estimate. Positive definite matrix $P_t$ is the solution to $\frac{dP}{dt} = A_tP_t+P_tA_t^T+Q_t-P_tc^TR_t^{-1}cP_t,$
where $Q_t=\sigma_t\sigma_t^T$, $A_t=\frac{\partial \Bar{f}}{\partial x}(\hat{x}_t,u_t)$, and $\Bar{f}(x,u)=f(x)+g(x)u$.
We make the following assumption in this paper.

\begin{assumption}
\label{assumption:ekf}
When $a_t=0$, the SDEs (\ref{eq:state-sde})-(\ref{eq:output-sde}) satisfy:
\begin{enumerate}
\item There exist constants $\beta_{1}$ and $\beta_{2}$ such that $\mathbf{E}(\sigma_{t}\sigma_{t}^{T}) \geq \beta_{1} I$ and $\mathbf{E}(\nu_{t}\nu_{t}^{T}) \geq \beta_{2} I$ for all $t$. 
\item The pair $[\frac{\partial \overline{f}}{\partial x}(x,u),c]$ is uniformly detectable.
\item Let $\phi$ be defined by $\overline{f}(x,u)-\overline{f}(\hat{x},u) = \frac{\partial \overline{f}}{\partial x}(x - \hat{x}) + \phi(x,\hat{x},u).$ Then there exist real numbers $k_{\phi}$ and $\epsilon_{\phi}$ such that $\|\phi(x,\hat{x},u)\| \leq k_{\phi}\|x-\hat{x}\|_{2}^{2}$ for all $x$ and $\hat{x}$ satisfying $\|x-\hat{x}\|_{2} \leq \epsilon_{\phi}$. 
\end{enumerate}
\end{assumption} 
The accuracy of of EKF is given by the theorem below.
\begin{theorem}[\cite{reif2000stochastic}]
\label{theorem:EKF}
Suppose that Assumption \ref{assumption:ekf} holds. Then there exists $\delta > 0$ such that $\sigma_{t}\sigma_{t}^{T} \leq \delta I$ and $\nu_{t}\nu_{t}^{T} \leq \delta I$. For any $0<\epsilon <1$, there exists $\gamma > 0$ such that $$Pr\left(\sup_{t \geq 0}{||x_{t}-\hat{x}_{t}||_{2} \leq \gamma}\right) \geq 1-\epsilon.$$ 
\end{theorem}
We finally introduce stochastic control barrier functions for robots in the absence of $a_t$, along with its safety guarantee.
\begin{theorem}[\cite{clark2020control}]
\label{theorem:SCBF}
    Suppose that $a_t=0$. For the robot in Eq. (\ref{eq:state-sde})-(\ref{eq:output-sde}) with safety region defined by Eq. (\ref{eq:safe-region}), define
    $$\overline{b}^{\gamma} = \sup_{x,x^{0}}{\{b(\hat{x}) : ||x-x^{0}||_{2} \leq \gamma \text{ and } b(x^{0})=0\}}. $$
    Let $\hat{b}^\gamma(\hat{x}) := b(\hat{x})-\overline{b}^{\gamma}$ and $\hat{x}_{t}$ denote the EKF estimate of $x_{t}$.
    Suppose that there exists a constant $\delta > 0$ such that   
    whenever $\hat{b}(\hat{x}_{t}) < \delta$, $u_{t}$ is chosen to satisfy
    \begin{multline}
    \label{eq:SCBF-2}
    \frac{\partial b}{\partial x}(\hat{x}_{t})\overline{f}(\hat{x}_{t},u_{t}) - \gamma\|\frac{\partial b}{\partial x}(\hat{x}_{t})K_{t}c\|_{2} \\
    + \frac{1}{2}\mathbf{tr}\left(\nu_{t}^{T}K_{t}^{T}\frac{\partial^{2}b}{\partial x^{2}}(\hat{x}_{t})K_{t}\nu_{t}\right) \geq -\hat{b}^\gamma(\hat{x}_{t}).
    \end{multline}
    Then $Pr(x_{t} \in \mathcal{D} \ \forall t | \ \|x_{t}-\hat{x}_{t}\|_{2} \leq \gamma \ \forall t) = 1$, where $\mathcal{D}=\{x\mid b(x)\geq 0\}$. 
\end{theorem}

The function $b$ satisfying inequality \eqref{eq:SCBF-2} is a stochastic control barrier function.
It specifies that as the state approaches the boundary, the control input is chosen such that the rate of increase of the barrier function decreases to zero. 
Hence Theorem \ref{theorem:SCBF} implies that if there exists a stochastic control barrier function for a system, then the safety condition is satisfied with probability $(1-\epsilon)$ when an EKF is used as an estimator and the control input is chosen to satisfy Eq. (\ref{eq:SCBF-2}).




\section{Solution Approach to Safety-Critical Control and Synthesis of Fault Tolerant NCBF}\label{sec:sol}

In this section, we first present an overview of our solution to safety-critical control synthesis for the robot in Eq. \eqref{eq:state-sde}-\eqref{eq:output-sde} such that safety can be guaranteed under sensor faults and attacks.
The key to our approach is the development of a new class of control barrier functions named \emph{fault tolerant neural control barrier functions (FT-NCBFs)}.

\subsection{Overview of Proposed Solution}

This subsection presents our proposed solution approach to safety-critical control synthesis.
Since the attack pattern is unknown, we maintain a set of $m$ EKFs, where each EKF uses measurements from $\{1,\ldots,q\}\setminus\mathcal{F}(r_i)$ for each $i\in\{1,\ldots,m\}$. 
We denote the state estimates and Kalman gain obtained using $\{1,\ldots,q\}\setminus\mathcal{F}(r_i)$ as $\hat{x}_{t,i}$ and $K_{t,i}$, respectively.
If there exists a function $b_\theta$ parameterized by $\theta$ such that $\mathcal{D}_\theta = \{\hat{x}|b_\theta(\hat{x})\geq 0\}\subseteq\mathcal{C}$, then Theorem \ref{theorem:SCBF} indicates that any control input $u$ within the feasible region
\begin{multline*}
    \Omega_i = \{u:\frac{\partial b_\theta}{\partial x}f(\hat{x}_{t,i}) + \frac{\partial b_{\theta}}{\partial x}g(\hat{x}_{t,i}) u -\gamma_i
    \|\frac{\partial b_\theta}{\partial x}(\hat{x})K_{t,i}c_i\|_{2}\\
    +  \frac{1}{2}\mathbf{tr}\left(\nu_i^{T}K_{t,i}^{T}\frac{\partial^{2}b_\theta}{\partial x^{2}}(\hat{x}_{t,i})K_{t,i}\nu_i\right)  +  \hat{b}^{\gamma_i}_\theta(\hat{x}_{t,i}) \geq 0\},
\end{multline*}
guarantees safety under attack pattern $r_i$,
where $c_i$ is obtained by removing rows corresponding to $\mathcal{F}(r_i)$ from matrix $c$, $\hat{b}_\theta^{\gamma_i}(\hat{x})=b_\theta(\hat{x})-\Bar{b}_\theta^{\gamma_i}(\hat{x})$, and
\begin{equation*}
\overline{b}_{\theta}^{\gamma_{i}}
= \sup_{\hat{x},\hat{x}^{0}}{\left\{b_{\theta}(\hat{x}) : ||\hat{x}-\hat{x}^{0}||_{2} \leq \gamma_{i}\right.} \\
\left. \text{ and } b_{\theta}(\hat{x}^{0})=0\right\}.
\end{equation*}
If there exists a control input $u\in\cap_{i=1}^m\Omega_i\neq\emptyset$, such a control input can guarantee the safety under any attack pattern $r_i$.

We note that the existence of a control input $u$ satisfying the constraints specified by $\Omega_1,\ldots,\Omega_m$ simultaneously may not be guaranteed because sensor faults and attacks can significantly bias the state estimates.
Thus we develop a mechanism to identify constraints conflicting with each other, and resolve such conflicts.
Our idea is to additionally maintain ${m \choose 2}$ EKFs, where each EKF computes state estimates using sensors from $\{1,\ldots,q\}\setminus(\mathcal{F}(r_i)\cup\mathcal{F}(r_j))$ for all $i\neq j$.
We use a variable $Z_t$ to keep track of the attack patterns that will not raise conflicts.
The variable $Z_t$ is initialized as $\{1,\ldots,m\}$.
If $\cap_{i\in Z_t}\Omega_i=\emptyset$, we compare state estimates $\hat{x}_{t,i}$ with $\hat{x}_{t,j}$ for all $i,j\in Z_t$ and $i\neq j$.
If $\|\hat{x}_{t,i}-\hat{x}_{t,j}\|_2\geq \alpha_{ij}$ for some chosen parameter $\alpha_{ij}>0$, then $Z_t$ is updated as 
\begin{equation*}
    Z_t = \begin{cases}
        Z_t\setminus \{i\},&\mbox{ if }\|\hat{x}_{t,i}-\hat{x}_{t,i,j}\|_2\geq \alpha_{ij}/2\\
        Z_t\setminus \{j\},&\mbox{ if }\|\hat{x}_{t,j}-\hat{x}_{t,i,j}\|_2\geq \alpha_{ij}/2
    \end{cases}.
\end{equation*}
After updating $Z_t$, if $\cap_{i\in Z_t}\Omega_i\neq\emptyset$, then control input $u_t$ can be chosen as
\begin{equation}\label{eq:control synthesis}
    \min_{u_t\in\cap_{i\in Z_t}\Omega_i}~u_t^Tu_t.
\end{equation}
Otherwise, we will remove indices $i$ corresponding to attack pattern $r_i$ causing largest residue $y_{t,i}-c_i\hat{x}_{t,i}$ until $\cap_{i\in Z_t}\Omega_i\neq \emptyset$.
Here $y_{t,i}$ is the output from sensors in $\{1,\ldots,q\}\setminus \mathcal{F}(r_i)$.

The positive invariance of set $\mathcal{D}_\theta$ using the procedure described above is established in the following theorem.
\begin{theorem}[\cite{clark2020control}]\label{thm:safety}
    Suppose  $\gamma_{1},\ldots,\gamma_{m}$, and $\alpha_{ij}$ for $i < j$ are chosen such that the following conditions are satisfied:
    \begin{enumerate}
    \item Define  $\Lambda_{i}(\hat{x}_{t,i}) = \frac{\partial b_\theta}{\partial x}g(\hat{x}_{t,i})$. There exists $\delta > 0$ such that for any $X_{t}^{\prime} \subseteq X_{t}(\delta):=\{i\mid \hat{b}_{\theta}^{\gamma_i}(\hat{x}_{t,i})<\delta\}$ satisfying $||\hat{x}_{t,i}-\hat{x}_{t,j}||_{2} \leq \alpha_{ij}$ for all $i,j \in X_{t}^{\prime}$,  there exists $u$ such that
    \begin{multline}
    \label{eq:def ftncbf}
    \Lambda_{i}(\hat{x}_{t,i})u > -\frac{\partial b_\theta}{\partial x}f(\hat{x}_{t,i}) +\gamma_i\|\frac{\partial b_\theta}{\partial x}(\hat{x}_{t,i})K_{t,i}c_i\|_{2}\\
    -  \frac{1}{2}\mathbf{tr}\left(\nu_i^{T}K_{t,i}^{T}\frac{\partial^{2}b_\theta}{\partial x^{2}}(\hat{x})K_{t,i}\nu_i\right)  -  \hat{b}^{\gamma_i}_\theta(\hat{x}_{t,i})
    \end{multline}
    for all $i \in X_t'$.
    \item For each $i$, when $r=r_{i}$, 
    \begin{multline}
    Pr(\|\hat{x}_{t,i}-\hat{x}_{t,i,j}\|_{2} \leq \alpha_{ij}/2 \ \forall j, \|\hat{x}_{t,i}-x_{t}\|_{2} \leq \gamma_{i} \ \forall t)\\
     \geq 1-\epsilon.
    \end{multline}
    \end{enumerate}
    Then $Pr(x_{t} \in \mathcal{D}_\theta \ \forall t) \geq 1-\epsilon$ for any $r \in \{r_{1},\ldots,r_{m}\}$.
\end{theorem}

Based on Theorem \ref{thm:safety}, we note that the key to our solution approach is to find the function $b_\theta$. 
We name the function $b_\theta$ as \emph{fault tolerant neural control barrier function (FT-NCBF)}, whose definition is given as below. 
\begin{definition}
    A function $b_\theta$ parameterized by $\theta$ is a fault tolerant neural control barrier function for the robot in Eq. \eqref{eq:state-sde}-\eqref{eq:output-sde} if it there exists a control input $u$ satisfying Eq. \eqref{eq:def ftncbf} under the conditions in Theorem \ref{thm:safety}.
\end{definition}

Solving Problem \ref{prob:1} hinges on the task of synthesizing an FT-NCBF for the robot in \eqref{eq:state-sde}-\eqref{eq:output-sde}, which will be our focus in the remainder of this section.
Specifically, we first investigate how to synthesize NCBFs when there exists no adversary (Section \ref{sec:SNCBF}).
We then use the NCBFs as a building block, and present how to synthesize FT-NCBFs.
We construct a loss function to learn FT-NCBFs in Section \ref{sec:FT-NCBF}.
We establish the safety guarantee of our approach in Section \ref{sec:conflict}.

\subsection{Synthesis of NCBF}\label{sec:SNCBF}


In this subsection, we describe how to synthesize NCBFs. 
We first present the necessary and sufficient conditions for stochastic control barrier functions, among which NCBFs constitute a special class represented by neural networks. 
\begin{proposition}
\label{prop:verify_scbf}
Suppose Assumption \ref{assumption:ekf} holds. The function $b(\hat{x})$ is a stochastic control barrier function if and only if there is no $\hat{x}\in \mathcal{D}^{\gamma}:=\{\hat{x}\mid \hat{b}(\hat{x})\geq 0\}$, satisfying
$\frac{\partial b}{\partial x}g(\hat{x}) = 0$ and $\xi^\gamma(\hat{x}) < 0$, where 
\begin{multline}
\label{eq:prop2cond}
    \xi^\gamma(\hat{x}) =  \frac{\partial b}{\partial x}f(\hat{x}) +  \frac{1}{2}\mathbf{tr}\left(\nu^{T}K_{t}^{T}\frac{\partial^{2}b}{\partial x^{2}}(\hat{x})K_{t}\nu\right)  \\
    -\gamma||\frac{\partial b}{\partial x}(\hat{x})K_{t}c||_{2} +  \hat{b}^\gamma(\hat{x}).
\end{multline}
\end{proposition}
The proposition is based on \cite[Proposition 2]{zhang2022safe}. We omit the proof due to space constraint. 
We note that the class of NCBFs is a special subset of stochastic control barrier functions.
We denote the NCBF as $b_\theta(\hat{x})$, where $\theta$ is the parameter of the neural network representing the function. 

In the following, we introduce the concept of \emph{valid} NCBFs, and present how to synthesize them. 
A valid NCBF needs to satisfy the following two properties.
\begin{definition}[Correct NCBFs]
    \label{def:correctness}
    Given a safety region $\mathcal{C}$, the NCBF $b_{\theta}$ is correct if and only if $\mathcal{D}_{\theta}\subseteq \mathcal{C}$. 
\end{definition}
The correctness property requires the NCBF $b_\theta$ to induce a set $\mathcal{D}_\theta\subseteq \mathcal{C}$.
If $\mathcal{D}_\theta$ is positive invariant, then $\mathcal{C}$ is also positive invariant, ensuring the robot to be safe with respect to $\mathcal{C}$.
We next give the second property of valid NCBFs.
\begin{definition}[Feasible NCBF]
\label{def:feasibility_scbf}
    The NCBF $b_\theta$ parameterized by $\theta$ is feasible if and only if $\forall \hat{x}\in \mathcal{D}_\theta^{\gamma}:=\{\hat{x}|\hat{b}_\theta^\gamma(\hat{x})\geq 0\}$, there exists $ u$ such that $\xi_\theta^\gamma(\hat{x}) + \frac{\partial b_\theta}{\partial x}g(\hat{x})u\geq 0$, where
    \begin{multline} \label{eq:prop2cond-2}
    \xi_\theta^\gamma(\hat{x}) =  \frac{\partial b_\theta}{\partial x}f(\hat{x}) +  \frac{1}{2}\mathbf{tr}\left(\nu^{T}K_{t}^{T}\frac{\partial^{2}b_\theta}{\partial x^{2}}(\hat{x})K_{t}\nu\right)  \\
    -\gamma\|\frac{\partial b_\theta}{\partial x}(\hat{x})K_{t}c\|_{2} +  \hat{b}_\theta^\gamma(\hat{x}).
\end{multline}
\end{definition}
The feasibility property in Definition \ref{def:feasibility_scbf} ensures that a control input $u$ can always be found to satisfy the inequality \eqref{eq:SCBF-2}, and hence can guarantee safety.


We note that there may exist infinitely many valid NCBFs. 
In this work, we focus on synthesizing valid NCBFs that encompass the largest possible safety region. To this end, we define an operator $Vol(\mathcal{D}_\theta)$ to represent the volume of the set $\mathcal{D}_\theta$, and synthesize a valid NCBF such that $Vol(\mathcal{D}_\theta)$ is maximized. The optimization program is given as follows
\begin{align}
    \label{eq:train_obj}
    \max_{\theta}&\quad{Vol(\mathcal{D}_\theta)} \\
    \label{eq:train_feasible_cons}
    s.t. & \quad \xi_\theta^\gamma(\hat{x}) \geq 0 \quad \forall \hat{x}\in \partial \mathcal{D}_\theta^{\gamma} \\
    \label{eq:train_correct_cons}
    & \quad b_{\theta}(\hat{x}) \leq h(\hat{x}) \quad \forall \hat{x}\in \mathcal{X}\backslash\mathcal{D}_\theta
\end{align}
where $\partial \mathcal{D}_\theta^{\gamma}$ represents the boundary of set $\mathcal{D}_\theta^{\gamma}$.
Here constraints \eqref{eq:train_feasible_cons} and \eqref{eq:train_correct_cons} require parameter $\theta$ to define feasible and correct NCBFs, respectively.
Solving the constrained optimization problem is challenging. In this work, we convert the constrained optimization to an unconstrained one by constructing a loss function which penalizes violations of the constraints.
We then minimize the loss function over a training dataset to learn parameters $\theta$ and thus NCBF $b_\theta$.

We denote the training dataset as $\mathcal{T}:=\{\hat{x}_{1}, \ldots, \hat{x}_{N}\mid \hat{x}_{i}
\in \mathcal{X}, \forall i =1,\ldots,N\}$, where $N$ is the number of samples. The dataset $\mathcal{T}$ is generated by simulating estimates with fixed point sampling as in \cite{dawson2023safe}. We first uniformly discretize the state space into cells with length vector $L$. Next, we uniformly sample the center of discretized cell as fixed points $x_{f}$. Then we simulate the estimates by introducing a perturbation $\rho[j]$  sampled uniformly from interval $[x_{f}[j]-0.5 L[j], x_{f}[j]+0.5 L[j]]$. Finally, we have the sampling data $\hat{x}_i=x_{f}+\rho\in \mathcal{T}\subseteq \mathcal{X}$.

We then formulate the following unconstrained optimization problem to search for $\theta$
\begin{equation}
    \label{eq:uncons_opt}
    \min_{\theta}\quad{-Vol(\mathcal{D}_{\theta})} + \lambda_{f}\mathcal{L}_f(\mathcal{T}) + \lambda_{c}\mathcal{L}_c(\mathcal{T}) 
\end{equation}
where $\mathcal{L}_f(\mathcal{T})$ is the loss penalizing the violations of constraint \eqref{eq:train_feasible_cons}, $\mathcal{L}_c(\mathcal{T})$ penalizes the violations of constraint \eqref{eq:train_correct_cons}, and $\lambda_{f}$ and $\lambda_{c}$ are non-negative coefficients. 
The objective function \eqref{eq:train_obj} is approximated by the following quantity
\begin{equation}
    \label{eq:volume_obj}
    Vol(\mathcal{D}_{\theta}) = \sum_{\hat{x}\in \mathcal{T}}{-ReLU(h(\hat{x}))ReLU(-b_{\theta}(\hat{x}))}. 
\end{equation}
Eq. \eqref{eq:volume_obj} penalizes the samples $\hat{x}$ in the safety region but not in $\mathcal{D}_\theta$, i.e., $h(\hat{x})>0$ and $b_\theta(\hat{x})<0$. 
The penalty of violating the feasibility property in Eq. \eqref{eq:train_feasible_cons} is defined as
\begin{equation*}\label{eq:Lf-SNCBF}
    \mathcal{L}_f(\mathcal{T}) = \sum_{\hat{x}\in \mathcal{T}} -\Delta(\hat{x}) ReLU(-\xi_\theta^\gamma(\hat{x}) - \frac{\partial b_{\theta}}{\partial x}g(\hat{x}) u + \hat{b}^\gamma_\theta(\hat{x})), 
\end{equation*}
where $\Delta(\hat{x})$ is an indicator function such that $\Delta(\hat{x}):= 1$ if $b_{\theta}(\hat{x})=\overline{b}^{\gamma}_\theta$ and $\Delta(\hat{x}):= 0$ otherwise. The function $\Delta$ allows us to find and penalize sample points $\hat{x}$ satisfying $\hat{b}_{\theta}^\gamma(\hat{x})=0$ and  $\xi_\theta^\gamma(\hat{x}) + \frac{\partial b_{\theta}}{\partial x}g(\hat{x}) u <0 $.
For each sample $\hat{x}\in\mathcal{T}$, the control input $u$ in $\mathcal{L}_f$ is computed as follows 
\begin{equation}
\begin{split}
    \min_{u} & \quad u^{T}u \\
    s.t. & \quad \xi_\theta^\gamma(\hat{x}) + \frac{\partial b_{\theta}}{\partial x}g(\hat{x}) u \geq 0
\end{split}
\end{equation}
The loss function to penalize the violations of the correctness property in Eq. \eqref{eq:train_correct_cons} is constructed as 
\begin{equation}
    \label{eq:correctness_loss}
    \mathcal{L}_c(\mathcal{T}) = \sum_{\hat{x}\in \mathcal{T}}{ReLU(-h(\hat{x}))ReLU(b_{\theta}(\hat{x}))}
\end{equation}
Eq. \eqref{eq:correctness_loss} penalizes $\hat{x}$ outside the safety region but being regarded safe, i.e., $h(\hat{x})<0$ and $b_{\theta}(\hat{x})>0$. 
When $\mathcal{L}_c(\mathcal{T})$ and $\mathcal{L}_f(\mathcal{T})$ converge to $0$, constraints \eqref{eq:train_feasible_cons}-\eqref{eq:train_correct_cons} are satisfied.

\subsection{Synthesis of FT-NCBF}\label{sec:FT-NCBF}
In Section \ref{sec:SNCBF}, we presented the training of NCBFs when there exists no adversary.
In this subsection, we generalize the construction of the loss function in Eq. \eqref{eq:uncons_opt}, and present how to train a valid FT-NCBF for robotic systems in Eq. \eqref{eq:state-sde}-\eqref{eq:output-sde} under unknown attack patterns.
With a slight abuse of notations, we use $b_\theta$ to denote the FT-NCBF in the remainder of this paper. 
We define $\hat{b}_\theta^{\gamma_i}(\hat{x})=b_\theta(\hat{x})-\Bar{b}_\theta^{\gamma_i}(\hat{x})$, where
\begin{equation*}
\overline{b}_{\theta}^{\gamma_{i}}
= \sup_{\hat{x},\hat{x}^{0}}{\left\{b_{\theta}(\hat{x}) : ||\hat{x}-\hat{x}^{0}||_{2} \leq \gamma_{i}\right.} \\
\left. \text{ and } b_{\theta}(\hat{x}^{0})=0\right\}.
\end{equation*}
The following proposition gives the  necessary and sufficient conditions for a function $b_{\theta}$ to be an FT-NCBF.  

\begin{proposition}
\label{prop:verify_ftscbf}
Suppose Assumption \ref{assumption:ekf} holds. The function $b_{\theta}(\hat{x})$ is an FT-NCBF if and only if there is no $\hat{x}_{t,i}\in \mathcal{D}_\theta^{\gamma_i}$, satisfying
$\frac{\partial b_\theta}{\partial x}g(\hat{x}_{t,i}) = 0$ , $\xi_\theta^{\gamma_i}(\hat{x}_{t,i}) < 0$ for all $i\in \{1,\ldots,m\}$ where 
\begin{multline}
\label{eq:ftsncbf_cond}
    \xi_\theta^{\gamma_i}(\hat{x}_{t,i}) =  \frac{\partial b_\theta}{\partial x}f(\hat{x}_{t,i}) +  \frac{1}{2}\mathbf{tr}\left(\nu_i^{T}K_{t,i}^{T}\frac{\partial^{2}b_\theta}{\partial x^{2}}(\hat{x})K_{t,i}\nu_i\right)  \\
    -\gamma_i\|\frac{\partial b_\theta}{\partial x}(\hat{x}_{t,i})K_{t,i}c_i\|_{2} +  \hat{b}^{\gamma_i}_\theta(\hat{x}_{t,i}).
\end{multline}
\end{proposition}
The proposition can be proved using the similar idea to Proposition \ref{prop:verify_scbf}. We omit the proof due to space constraint.

We construct the loss function below to learn FT-NCBFs 
\begin{equation}
    \label{eq:train_ftsncbf}
    \min_{\theta}\quad{-Vol(\mathcal{D}_{\theta})} + \lambda_f\sum_{i\in \{1,\ldots,m\}}{\mathcal{L}^{i}_{f}}(\mathcal{T}) + \lambda_c\mathcal{L}_c(\mathcal{T}), 
\end{equation}
where $\mathcal{L}_{f}(\mathcal{T}) = \sum_{i\in \{1,\ldots,m\}}{\mathcal{L}^{i}_{f}}(\mathcal{T})$ is the penalty of violating the feasibility property,
\begin{equation*}
    \mathcal{L}^{i}_{f}(\mathcal{T}) = \sum_{\hat{x}\in \mathcal{T}} -\overline{\Delta}_{i}(\hat{x}) ReLU(-\xi_\theta^{\gamma_i}(\hat{x}) - \frac{\partial b_{\theta}}{\partial x}g(\hat{x}) u + \hat{b}_\theta^{\gamma_i}(\hat{x})), 
\end{equation*}
and $\overline{\Delta}(\hat{x})$ is an indicator function such that $\overline{\Delta}(\hat{x}):= 1$ if $b_{\theta}(\hat{x})\leq\max_{i\in Z_t}{\{\overline{b}^{\gamma_i}_\theta\}}$ and $\overline{\Delta}(\hat{x}):= 0$ otherwise. 
The control input $u$ used to compute $\mathcal{L}_f^i(\mathcal{T})$ for each sample $\hat{x}$ is calculated as follows. 
\begin{equation}
\label{eq:ctrl_multiSNCBF}
\begin{split}
    \min_{u} & \quad u^{T}u \\
    s.t. & \quad \xi_\theta^{\gamma_i}(\hat{x}) + \frac{\partial b_{\theta}}{\partial x}g(\hat{x}) u \geq 0 \quad \forall i\in \{1,\ldots,m\}
\end{split}
\end{equation}
If $\mathcal{L}_c(\mathcal{T})$ and $\mathcal{L}_f(\mathcal{T})$ converge to $0$, $b_{\theta}$ is a valid FT-NCBF.

\subsection{Safety Guarantee of Proposed Approach}\label{sec:conflict}


In this subsection, we establish the safety guarantee of our approach for the robot in Eq. \eqref{eq:state-sde}-\eqref{eq:output-sde}.
First, we note that Theorem \ref{thm:safety} establishes the positive invariance of set $\mathcal{D}_\theta$.
However, the theorem depends on the existence of $u_{t}$. 
The following proposition provides the sufficient condition of the existence of $u_{t}$ for all $\hat{x}\in \mathcal{D}^{\gamma_i}_{\theta}, \ \forall i\in Z_{t}$.

\begin{proposition}
    \label{prop:u_f_ftsncbf}
    Suppose that the interval length $L$ used to sample the training dataset $\mathcal{T}$ satisfies $L\leq s$ and $s\rightarrow0$. If an FT-NCBF $b_{\theta}$ satisfies  $\mathcal{L}_f(\mathcal{T})+\mathcal{L}_c(\mathcal{T})= 0$,
    then there always exists $u_{t}$ such that $\frac{\partial b_{\theta}}{\partial x}g(\hat{x})u_{t}+\xi_\theta^{\gamma_i}(\hat{x})\geq 0$ $\forall \hat{x}\in \mathcal{D}_{\theta}^{\gamma_i}, \ \forall i\in Z_t$. 
\end{proposition}
\begin{proof}
    By the constructions of $\mathcal{L}_f^i$ and $\mathcal{L}_c$, these losses are non-negative. Thus if $\mathcal{L}_f(\mathcal{T})+\mathcal{L}_c(\mathcal{T})= 0$, we have $\mathcal{L}_f^i(\mathcal{T})=\mathcal{L}_f(\mathcal{T})=\mathcal{L}_c= 0$.
    According to the definitions of $\mathcal{L}_f$ and $\mathcal{L}_f^i$ as well as the conditions that $L\leq s$ and $s\rightarrow0$, we then have that there must exist some control input $u$ that solves the optimization program in Eq. \eqref{eq:ctrl_multiSNCBF} for all $ \hat{x}\in \mathcal{D}_{\theta}^{\gamma_i}$ when  $\mathcal{L}_f^i(\mathcal{T})+\mathcal{L}_f(\mathcal{T})=0$. Otherwise losses $\mathcal{L}_f^i$ and $\mathcal{L}_f$ will be positive.
\end{proof}
We finally present the safety guarantee of our approach. 
\begin{theorem}
    \label{th:FT_Safety}
    Suppose that the interval length $L$ used to sample the training dataset $\mathcal{T}$ satisfies $L\leq s$ and $s\rightarrow0$.
    Let $b_\theta$ be an FT-NCBF satisfying $\mathcal{L}_f(\mathcal{T})+\mathcal{L}_c(\mathcal{T})= 0$.
    Suppose $\gamma_{1},\ldots,\gamma_{m}$, and $\alpha_{ij}$ for $i < j$ are chosen such that the conditions in Theorem \ref{thm:safety} hold. Then $Pr(x_{t} \in \mathcal{C} \ \forall t) \geq 1-\epsilon$ for any attack pattern $r \in \{r_{1},\ldots,r_{m}\}$. 
\end{theorem}
\begin{proof}
    The theorem follows from Theorem \ref{thm:safety}, Proposition \ref{prop:u_f_ftsncbf}, and the correctness property that $\mathcal{D}_\theta\subseteq\mathcal{C}$.
\end{proof}

\begin{figure*}[htp]
\begin{subfigure}{.26\textwidth}
    \centering
    \includegraphics[width=\textwidth]{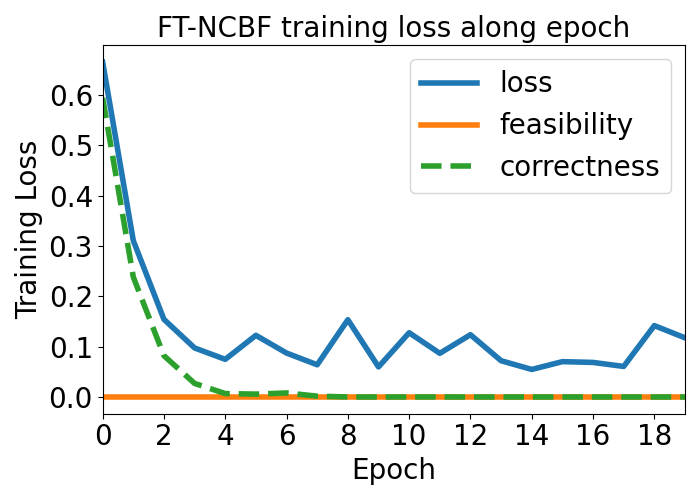}
    \subcaption{}
    \label{fig:training_curve}
\end{subfigure}%
\hfill
\begin{subfigure}{.26\textwidth}
    \centering
    \includegraphics[width=\textwidth]{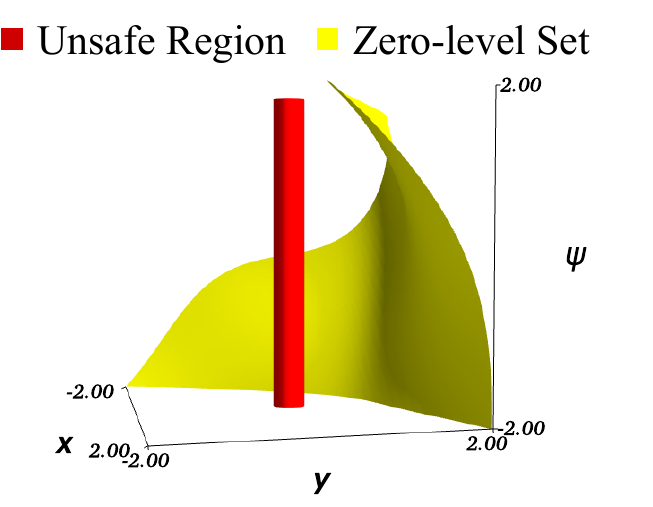}
    \subcaption{}
    \label{fig:obs0level}
\end{subfigure}%
\hfill
\begin{subfigure}{.39\textwidth}
    \centering
    \includegraphics[width=\textwidth]{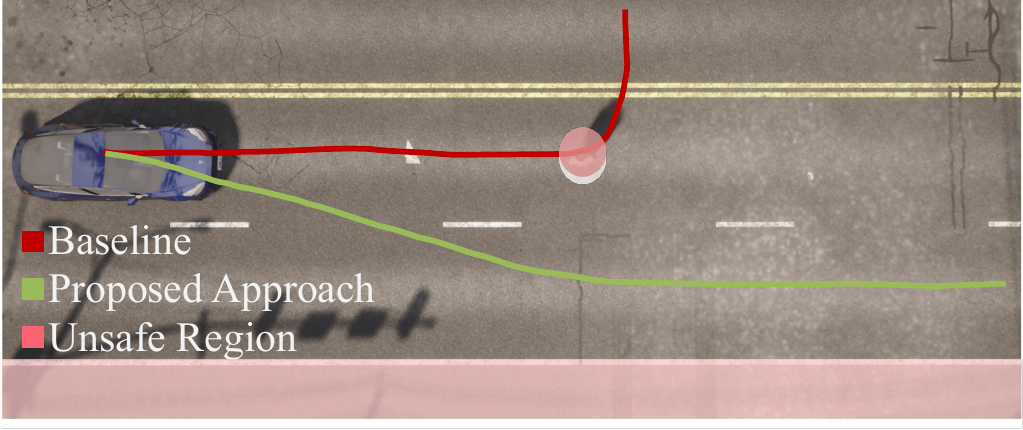}
    \subcaption{}
    \label{fig:carla_traj}
\end{subfigure}
\caption{This figure presents the experimental results on obstacle avoidance of an autonomous mobile robot.
Fig. \ref{fig:training_curve} presents the values of loss function, $\mathcal{L}_f(\mathcal{T})$, and $\mathcal{L}_c(\mathcal{T})$. The loss function decreases towards zero  during the training process.
Fig. \ref{fig:obs0level} shows the zero-level set of $\mathcal{D}_\theta$ corresponding to the FT-NCBF $b_\theta$. 
The set $\mathcal{D}_\theta$ does not overlap with the unsafe region in red color.
Fig. \ref{fig:carla_traj} presents the trajectory of the mobile robot when using control policies obtained by our approach and the baseline approach. 
We observe that our approach guarantees safety whereas the baseline crashes with the pedestrian.
}
\end{figure*}
\begin{figure*}[htp]
\begin{subfigure}{.26\textwidth}
    \centering
    \includegraphics[width=\textwidth]{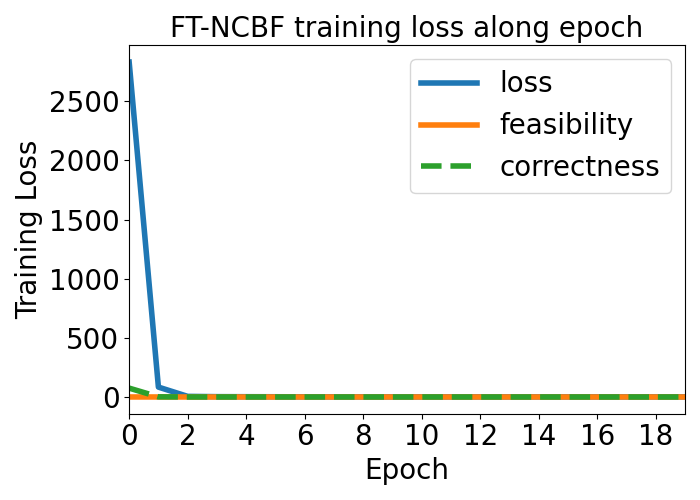}
    \subcaption{}
    \label{fig:lstraining_curve}
\end{subfigure}%
\hfill
\begin{subfigure}{.26\textwidth}
    \centering
    \includegraphics[width=\textwidth]{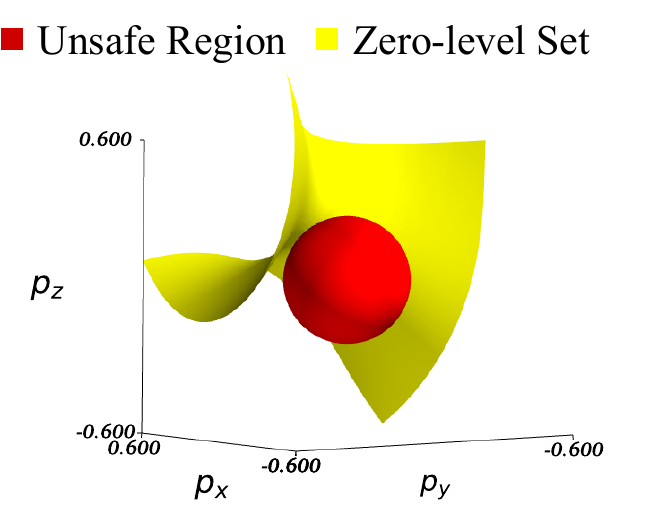}
    \subcaption{}
    \label{fig:ls0level}
\end{subfigure}%
\hfill
\begin{subfigure}{.38\textwidth}
    \centering
    \includegraphics[width=\textwidth]{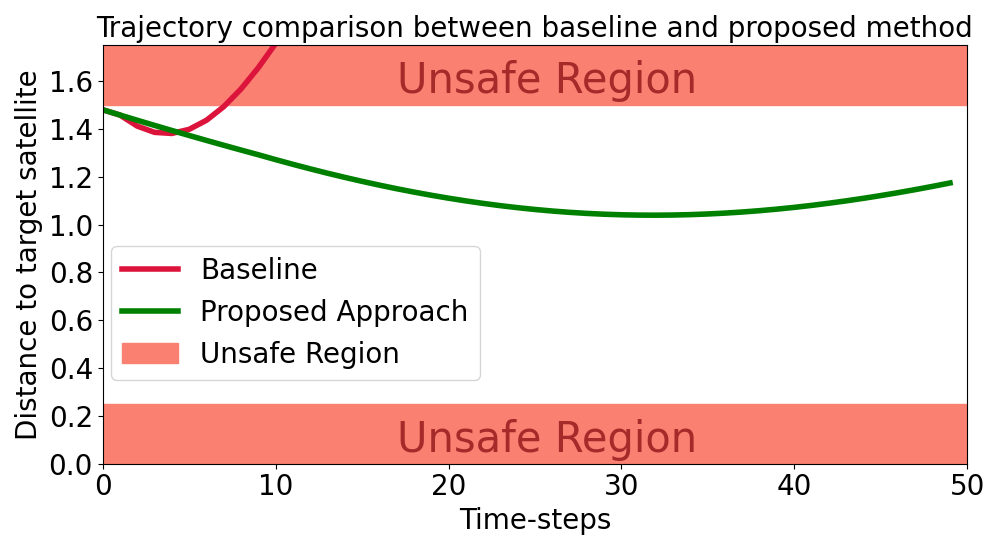}
    \subcaption{}
    \label{fig:dis_traj}
\end{subfigure}
\caption{This figure presents the experimental results on spacecraft rendezvous problem. In Fig. \ref{fig:lstraining_curve}, we demonstrate that the value of loss function in Eq. \eqref{eq:uncons_opt} quickly converges to zero during training. 
Fig. \ref{fig:ls0level} presents the zero-level set of $\mathcal{D}_\theta$, which never overlaps with the unsafe region in red color. 
Fig. \ref{fig:dis_traj} simulates the trajectories of the chaser satellite using our approach and the baseline.
We observe that our approach allows the chaser satellite to maintain a proper distance to the target satellite (green curve), whereas the baseline fails (red curve).}
\end{figure*}

\section{Experiments}\label{sec:exp}

In this section, we evaluate our proposed approach using two case studies, namely the obstacle avoidance of an autonomous mobile robot \cite{barry2012safety} and the spacecraft rendezvous problem \cite{jewison2016spacecraft}.
Both case studies are conducted on a laptop with an AMD Ryzen 5800H CPU and 32GB RAM. The hyper-parameters in both studies can be found in our code.


\subsection{Obstacle Avoidance Problem of Mobile Robot}\label{sec:obstacle}
We consider an autonomous mobile robot navigating on a road following the dynamics \cite{dubins1957curves} given below
\begin{equation*}
    \dot{x} = f(x) + g(x)u,
\end{equation*}
where $x:=[x_1, x_2, \psi]^T\in\mathcal{X}\subseteq\mathbb{R}^3$ is the state consisting of the location $(x_1,x_2)$ of the robot and its orientation $\psi$, $u$ is the input that controls the robot's orientation, $f(x) = [\sin\psi,\cos\psi,0]^T$, and $g(x)=[0,0,1]^T$.

The mobile robot is required to stay in the road while avoid pedestrians sharing the field of activities. 
We set the location of the pedestrian as $(0,0)$.
Then the safety region is formulated as $\mathcal{C} = \left\{\mathbf{x} \in \mathcal{X}: x_1^2+x_2^2 \geq 0.04, \text{ and } x_2 \geq -0.3\right\}$, where $\mathcal{X}= [-2,2]^3$.
We consider that one IMU and two GNSS sensors are mounted on the mobile robot.
These sensors jointly yield the output model $y = [x_1, x_1, x_2, x_2, \psi]^T + \nu$, where the measurement noise $\nu\sim \mathcal{N}(0,\sigma I_5)\in \mathbb{R}^5$, $\sigma = 0.001$, and $I_5$ is the five-dimensional identity matrix.
There exists an adversary who can spoof the readings from one GNSS sensor, leading to two possible attack patterns, $\{r_1,r_2\}$. 
The compromised sensors associated with attack patterns $r_1$ and $r_2$ are the second or fourth dimension of $y$, denoted as $y[2]$ and $y[4]$, respectively.

We compare our approach with a baseline which adopts the method from \cite{dawson2023safe} and learns an NCBF ignoring the presence of sensor faults and attacks.
The baseline computes the control input by solving $\min_{u\in\Bar{\Omega}}u^Tu$, where $\Bar{\Omega}$ is the feasible region specified by the learned NCBF.

When applying our approach, we first sample the training dataset $\mathcal{T}$ with $L=0.125$, making $|\mathcal{T}|=32^3$. 
Given the training dataset $\mathcal{T}$, we learn an FT-NCBF using Eq. \eqref{eq:train_ftsncbf} with $\gamma_1=0.002$ and $\gamma_2= 0.0015$.
The training process took about $604$ seconds.
The values of loss function, $L_f(\mathcal{T})$, and $\mathcal{L}_c(\mathcal{T})$ at each epoch during training are presented in Fig. \ref{fig:training_curve}.
We observe that the loss function decreases towards zero during the training process. 
In particular, $L_f(\mathcal{T})+\mathcal{L}_c(\mathcal{T})\rightarrow 0$ as we train more epochs.
By the construction of the loss function, it indicates that our approach finds a valid FT-NCBF.
The positive invariant set $\mathcal{D}_\theta$ induced by the learned FT-NCBF is shown in Fig. \ref{fig:obs0level}.
We observe that the zero-level set $\partial\mathcal{D}_\theta$ in yellow color stays close to the boundary of the safety region, while it does not overlap with the unsafe region in red color.
We implement the control policy calculated using our approach and simulate the trajectory of the mobile robot using CARLA \cite{dosovitskiy2017carla}.
In Fig. \ref{fig:carla_traj}, we observe that our proposed approach with parameter $\alpha_{12}=0.1$ avoids any contact with the pedestrian while remain in the road (green color curve) and thus is safe, whereas the baseline approach (red color curve) crashes with the pedestrian and hence fails.
A video clip of our simulation is available as the supplement.


\subsection{Spacecraft Rendezvous Problem}


In this section, we demonstrate the proposed approach using the spacecraft rendezvous between a chaser and a target satellite.
We follow the setting in \cite{jewison2016spacecraft}, and represent the dynamics of the satellites using the linearized Clohessy–Wiltshire–Hill equations as follows
\begin{equation*}
    \dot{x}=
    \left[\begin{array}{c c c c c c}
    
     &I_{3} & & & \mathbf{0}_{3} & \\
    
    3n^2 &0 &0 &0 &2n &0 \\
    0 &0 &0 &-2n &0 &0 \\
    0 &0 &-n^2 &0 &0 &0 
    \end{array}\right]
    x
+ \left[\begin{array}{c}
    \mathbf{0}_3 \\ I_3 
\end{array}\right]
 u
\end{equation*}
where $x=[p_x, p_y, p_z, v_x, v_y, v_z]^T$ is the state of the chaser satellite, $u=[u_x, u_y, u_z]^T$ is the control input representing the chaser's acceleration, and $n=0.056$ represents the mean-motion of the target satellite.

We define the state space and safety region as $\mathcal{X}=[-2,2]^6$ and $\mathcal{C}=\{x: r\in[0.25,1.5], r= \sqrt{p_x^2+ p_y^2+ p_z^2} \}$, respectively. The chaser satellite is required to maintain a safe distance from the target satellite as a safety constraint.
The chaser satellite is equipped with a set of sensors to obtain the output $y = [p_x, p_x, p_y, p_y, p_y, v_x, v_y, v_z]^T + \nu$, where $\nu\sim \mathcal{N}(0,\Sigma)$ and $\Sigma = 10^{-5}\times Diag([100, 100, 100, 1, 1, 1, 1, 1])$.
We consider two fault patterns $\{r_1,r_2\}$, where $r_1$ and $r_2$ are associated with compromised measurements from $y[2]$ and $y[4]$, respectively, raised by a perturbation $a\sim \mathcal{N}(-1,0.1)$.


We evaluate our approach by comparing with the same baseline approach in Section \ref{sec:obstacle}. 
We sample from state space $\mathcal{X}$ using $L=1$ and obtain a training dataset with $|\mathcal{T}|=4096$.
The training of FT-NCBF took about $1411$ seconds with the loss $\mathcal{L}_f(\mathcal{T})$, and $\mathcal{L}_c(\mathcal{T})$ shown in Fig. \ref{fig:lstraining_curve}.
We observe that the loss $\mathcal{L}_f(\mathcal{T})$ and $\mathcal{L}_c(\mathcal{T})$ quickly converge to $0$, and thus the learned FT-NCBF is valid.
We visualize the FT-NCBF $b_{\theta}$ in Fig. \ref{fig:ls0level}. 
We synthesize a control policy using $b_{\theta}$ in Eq. \eqref{eq:control synthesis}.
We observe in Fig. \ref{fig:dis_traj} that the chaser satellite never leaves the safety region using the control policy obtained by our approach, whereas the baseline fails to maintain a proper distance from the target satellite, leading to failures in the docking operation.


\section{Conclusion}\label{sec:conclusion}

In this paper, we focused on the problem of ensuring safety constraints for stochastic robotic systems under sensor faults and attacks.
To tackle the problem, we proposed FT-NCBFs and studied the synthesis of FT-NCBFs by first deriving the necessary and sufficient conditions for FT-NCBFs to guarantee safety.
We then developed a data-driven method to learn FT-NCBFs by minimizing a loss function which penalizes the violations of our derived conditions.
We investigated the safety-critical control synthesis using the learned FT-NCBFs and established the safety guarantee.
Specifically, we maintained a bank of EKFs to estimate system states, and developed a mechanism to resolve conflicting estimates raised by sensor faults and attacks. 
We demonstrated our approach using the obstacle avoidance of a mobile robot and spacecraft rendezvous. Future work will investigate practical limitations, including on-board computational complexity and data-driven dynamical models. 

\section*{Acknowledgment}
This research was supported by the AFOSR (grants FA9550-22-1-0054 and FA9550-23-1-0208), and NSF (grant CNS-1941670).

\bibliographystyle{ieeetr}
\bibliography{ref}
\end{document}